\documentclass[conference]{IEEEtran}
\usepackage{blindtext, graphicx}
\usepackage[hidelinks]{hyperref}
\usepackage{comment}
\usepackage{array}
\usepackage{url}
\usepackage{listings}
\usepackage{flushend}
\usepackage{multirow}
\usepackage{subfig}
\usepackage{color}
\usepackage{booktabs}

\usepackage{amsfonts}
\usepackage[boxed,ruled,vlined]{algorithm2e}

\usepackage{cite}
\ifCLASSINFOpdf
\else
\fi
\usepackage[cmex10]{amsmath}
\usepackage{algorithmic}

\begin{document}

% paper title
% can use linebreaks \\ within to get better formatting as desired

\title{Neighbourhood NILM: A Big-data Approach to Household Energy Disaggregation}
%\title{Using a dumb power meter for energy disaggregation }

% author names and affiliations
% use a multiple column layout for up to three different
% affiliations
\author{\IEEEauthorblockN{Nipun Batra}
\IEEEauthorblockA{ IIIT Delhi\\
	%Delhi, India\\
	nipunb@iiitd.ac.in}
\and
\IEEEauthorblockN{Amarjeet Singh}
\IEEEauthorblockA{ IIIT Delhi\\
	%Delhi, India\\
	amarjeet@iiitd.ac.in}

\and
\IEEEauthorblockN{\small Kamin Whitehouse}
\IEEEauthorblockA{ Univeristy of Virginia\\
	%Delhi, India\\
	whitehouse@virginia.edu}

}

% conference papers do not typically use \thanks and this command
% is locked out in conference mode. If really needed, such as for
% the acknowledgment of grants, issue a \IEEEoverridecommandlockouts
% after \documentclass

% for over three affiliations, or if they all won't fit within the width
% of the page, use this alternative format:
% 
%\author{\IEEEauthorblockN{Michael Shell\IEEEauthorrefmark{1},
%Homer Simpson\IEEEauthorrefmark{2},
%James Kirk\IEEEauthorrefmark{3}, 
%Montgomery Scott\IEEEauthorrefmark{3} and
%Eldon Tyrell\IEEEauthorrefmark{4}}
%\IEEEauthorblockA{\IEEEauthorrefmark{1}School of Electrical and Computer Engineering\\
%Georgia Institute of Technology,
%Atlanta, Georgia 30332--0250\\ Email: see http://www.michaelshell.org/contact.html}
%\IEEEauthorblockA{\IEEEauthorrefmark{2}Twentieth Century Fox, Springfield, USA\\
%Email: homer@thesimpsons.com}
%\IEEEauthorblockA{\IEEEauthorrefmark{3}Starfleet Academy, San Francisco, California 96678-2391\\
%Telephone: (800) 555--1212, Fax: (888) 555--1212}
%\IEEEauthorblockA{\IEEEauthorrefmark{4}Tyrell Inc., 123 Replicant Street, Los Angeles, California 90210--4321}}

% use for special paper notices
%\IEEEspecialpapernotice{(Invited Paper)}

% make the title area

\maketitle

\begin{abstract}
In this paper, we investigate whether ``big-data" is more valuable than ``precise" data for the problem of {\em energy disaggregation}: the process of breaking down aggregate energy usage on a per-appliance basis. Existing techniques for disaggregation rely on energy metering at a resolution of 1 minute or higher, but most power meters today only provide a reading once per month, and at most once every  15 minutes. In this paper, we propose a new technique called {\em
Neighborhood NILM} that leverages data from `neighbouring' homes to disaggregate energy given only a single energy reading per month. The key intuition behind our approach is that `similar' homes have `similar' energy consumption on a per-appliance basis. Neighborhood NILM matches every home with a set of `neighbours' that have direct submetering infrastructure, i.e. power meters on individual circuits or loads. Many such homes already exist. Then, it estimates the appliance-level energy consumption of the target home to be the average of its $K$ neighbours. We evaluate this approach using 25 homes and results show that our approach gives comparable or better disaggregation in comparison to state-of-the-art accuracy reported in the literature that depend on manual model training, high frequency power metering, or both. Results show that Neighbourhood NILM can achieve 83\% and 79\% accuracy disaggregating fridge and heating/cooling loads, compared to 74\% and 73\% for a technique called FHMM. Furthermore, it achieves up to 64\% accuracy on washing machine, dryer, dishwasher, and lighting loads, which is higher than previously reported results. Many existing techniques are not able to disaggregate these loads at all. These results indicate a potentially substantial advantage to installing submetering infrastructure in a select few homes rather than installing new high-frequency smart metering infrastructure in all homes.
\end{abstract}

% For peer review papers, you can put extra information on the cover
% page as needed:
% \ifCLASSOPTIONpeerreview
% \begin{center} \bfseries EDICS Category: 3-BBND \end{center}
% \fi
%
% For peerreview papers, this IEEEtran command inserts a page break and
% creates the second title. It will be ignored for other modes.
\IEEEpeerreviewmaketitle

\section{Introduction}

``Data is the new oil", so said Ann Winbland, who has been called one most influential people in the digital age. In this paper, we investigate whether ``big-data" is more valuable than ``precise" data for the problem of energy disaggregation.

{\em Energy disaggregation} is the process of breaking down the household aggregate energy measured at a single point into constituent appliances, akin the itemised bills we get from grocery stores. The process typically involves measuring household aggregate power using a smart meter, and using a model to break down the aggregate signal. Disaggregated energy can be used to save energy is several ways. For example, utility companies can direct energy
conservation programs toward specific homes that have unusually high energy
usage for certain appliances. Additionally, energy feedback can help people save
energy by creating behavior change. A recent pilot study in 850 homes performed
by the power utility PG\&E and startup company Bidgely found that eco-feedback
including a breakdown of household energy usage helped reduce energy usage by
7.7\% on average\footnote{http://www.bidgely.com/blog/pge-pilot-yields-7-7-energy-savings/}.

However, the practical impact of energy disaggregation is limited by three key
challenges. First, low-power loads such as lights get lost in the noise of
larger loads, even though lighting is the third largest energy consumer in the
average home~\cite{eia}. Second, all existing approaches to disaggregation require a model
of each electrical load, and a manual process is required to generate or train
these models~\cite{hart_1992, kolter_2012}. Techniques to learn these models automatically are
being developed but still have challenges with small loads such as lights and
complex loads such as washing machines and dishwashers~\cite{barker_2013}. Lastly and most importantly,
existing disaggregation techniques require power metering with approximately 1 minute
sampling intervals~\cite{parson_2012} and some require rates of 10 kHz or
more~\cite{berges2010enhancing, armel_2013}. However, the advanced metering infrastructure (AMI) that is currently being rolled out around the world provide data with 15-minute
or hourly smart meter reads, which is just enough for time-of-use
pricing. Companies are developing new technology to extract this information at
higher sampling rates\footnote{http://bit.ly/1LxZDCa}, but the impact of energy
disaggregation will grow only linearly with the rate at which this new
technology is rolled out.

\begin{figure*}[t]
	\centering
	\includegraphics[scale=1]{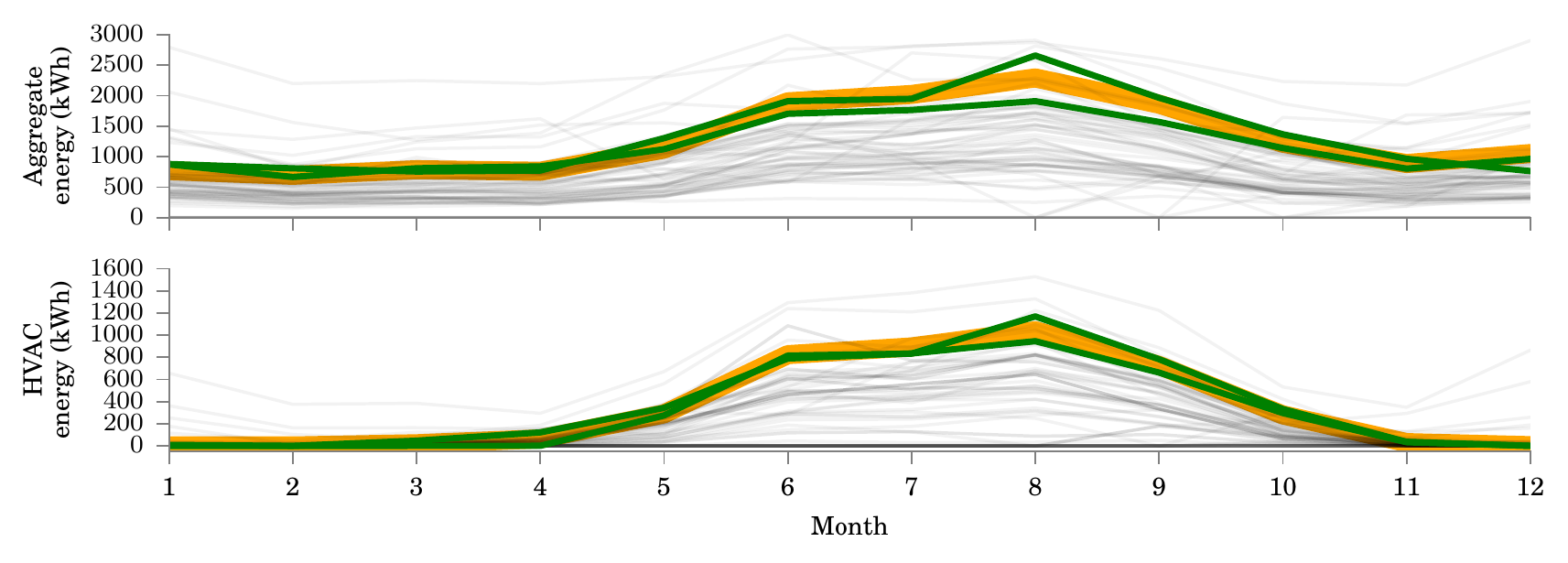} 
	
	\caption{Home similar in our feature set (aggregate energy consumption over 12 months, area, \# occupants, etc.) have similar energy consumption on a per-appliance basis. Orange line shows a test home and the green lines show its two nearest neighbours on our feature set. Black lines indicate other homes in the data set.}
	\label{fig:illustration} 
\end{figure*}

In this paper, we propose a new energy disaggregation technique called {\em
Neighborhood NILM} to overcome all three of these issues. This technique does not require high frequency data collected using smart meters. Instead, it leverages data from `neighbouring' homes to disaggregate energy given only a single energy reading per month. Neighbourhood NILM can be applied to any home that receives a monthly energy bill, and more frequent data collected through AMI meters could potentially make the approach more accurate. The key intuition behind our approach is that `similar' homes have `similar' energy consumption on a per-appliance basis. Neighborhood NILM requires three steps. First, we assume access to some number of homes with direct submetering infrastructure, i.e. power meters on individual circuits or loads. Many companies are already collecting submetering information from thousands of homes around the world\footnote{http://powerhousedynamics.com/}. Second, we match a new target home with its $K$ nearest neighbours from among the set of submetered homes based on its monthly energy bills, as well as static characteristics of a household such as the size of the home and the number of occupants. These characteristics are often publicly available and are already being used by companies to match homes based on similarity\footnote{https://opower.com/}. A home must only match with a small handful of homes in the submetered set in order to perform energy disaggregation. Finally, we estimate the appliance-level energy consumption of the target home to be the average appliance-level energy consumption across its $K$ neighbours.

Unlike conventional energy disaggregation, the potential of Neighbourhood NILM grows superlinearly as submetering infrastructure is rolled out. Indeed, Neighborhood NILM could probably be provided as a service today by the many companies that already have access to submetering data. Additionally, it does not require any models of power loads to be built or trained. Neighbourhood NILM only needs to decide which features indicate that two homes will be ``similar" in terms of appliance-level energy usage, but it determines these features automatically based on the submetered homes, whose true similarity is already known. Finally, Neighbourhood NILM functions even with low-power loads like lights and complex loads like washing machines and dishwashers. Any loads that are directly submetered on the neighboring homes will be disaggregated on the target home.

We evaluate this approach using 25 homes from the publicly available Dataport data set~\cite{parson2015dataport} that have direct submetering infrastructure installed for one year or more. Results show that our approach gives comparable or better disaggregation performance across all appliances that we explored in comparison to four benchmarks, including state-of-the-art accuracy reported in the literature that depend on manual model training, high frequency power metering, or both. Results show that Neighbourhood NILM can achieve 83\% and 79\% accuracy disaggregating fridge and heating/cooling loads, compared to 74\% and 73\% for a technique called FHMM~\cite{kolter_2012}. Furthermore, it achieves up to 64\% accuracy on washing machine, dryer, dishwasher, and lighting loads, which is higher than previously reported results. Many existing techniques are not able to disaggregate these loads at all.

Results indicate that a ``big data" approach that leverages submetering data from neighboring homes compares favourably to collecting high resolution data from a smart meter. Thus, any companies or utilities that want to produce a per-appliance energy breakdown should consider installing submetering infrastructure in a select few homes rather than new high-frequency smart metering infrastucture in all homes.

\begin{figure*}[t]
	\centering
	\includegraphics[scale=0.6]{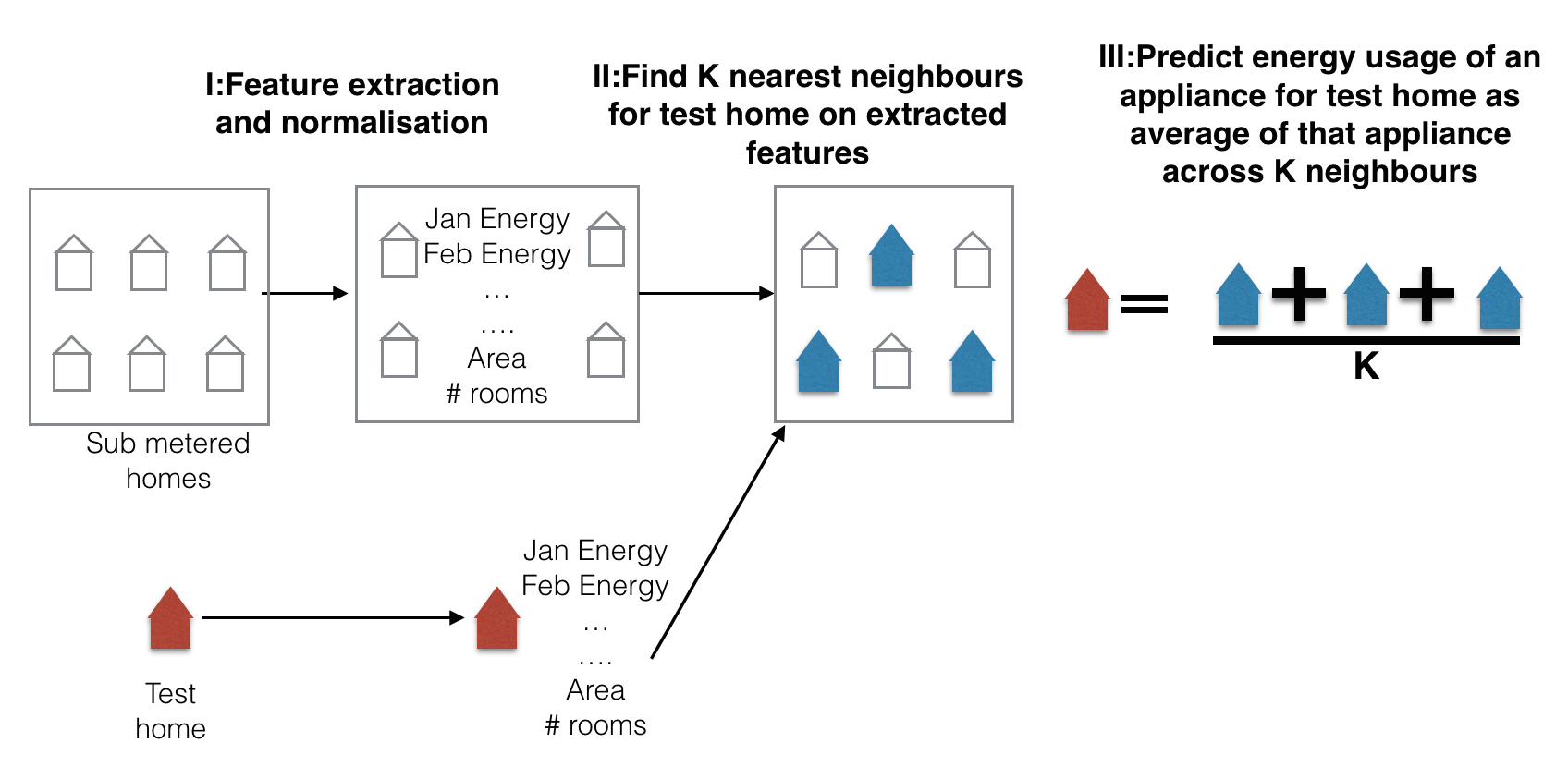} 
	
	\caption{Overview of Neighbourhood NILM. Neighbourhood NILM consists of three main steps: 1) Extracting and normalising features such as household historical monthly aggregate energy, area of home, for sub metered and test homes; 2) Finding $K$ nearest neighbours of the test home on these features in the set of submetered homes; and 3) Predicting the energy usage of an appliance in the test home as the average energy usage of that appliance across these $K$ neighbours}
	\label{fig:approach-overview} 
\end{figure*}

\section{Related Work}
There are three broad categories of related work in the literature: 1) traditional NILM approaches; 2) NILM approaches leveraging large quantities of data; and 3) studies on the inter-relationship between household characteristics and energy consumption. We describe each of these now.
\subsection{Traditional NILM approaches}
Since its inception in the early 1980s by George Hart~\cite{hart_1992}, the field of NILM has seen various approaches based on different machine learning techniques leveraging different features of the power trace. However, many of these approaches require submetered data to learn a model of each appliance and these models have not been shown to generalize well across homes. Even if the models did generalize, these approaches would require high frequency power metering with resolutions of 1-minute or higher. High frequency approaches (\textgreater 10 kHz)~\cite{berges2010enhancing} use features such as voltage-current trajectories to detect events in aggregate power time series. All of these techniques grow linearly with the roll out of submetering and smart metering infrastructure. Current smart meters do not collect data at such high rates because they are designed and deployed for the purposes of time-of-use pricing and there are currently no efforts to deploy devices suitable for energy disaggregation on a large scale. Therefore, these techniques while promising must face real practical barriers before going into effect.  

Additionally, existing approaches for energy disaggregation~\cite{kolter_2012, parson_2012, hart_1992} require a model of each appliance. 
The main differences between these techniques are how they are created and how they are used to infer the hidden states of the appliances based on the aggregate power readings. For example, some systems model appliances as finite state machines (FSMs). However, such approaches generally show poor accuracy on complex appliances such as washing machine and other electronics, as a FSM is a poor model for such appliances. Some systems assume the model is manually generated, learned from training data~\cite{hart_1992, kolter_2012}, and in rare cases learned automatically~\cite{barker_2013}. In all cases, however, the accuracy of these models depends on how well the model approximates the true appliances in the home and it has not yet been demonstrated that these model-based approaches generalize well across homes.

\subsection{NILM approaches leveraging large quantities of data}
In the recent past, large quantities of data, both in the number of homes and time duration have been made available. Recently, the availability of large data has shown a lot of promise in related machine learning fields such as image classification, using deep neural networks. Kelly et al.~\cite{kelly2015neural} take inspiration from the success of deep neural networks and apply it to the energy disaggregation problem, where they show that their technique outperforms various benchmarks across all appliances, including appliances which show poor accuracies when modelled as FSMs. Their hypothesis is that given ``enough" data, deep neural networks can learn features for an arbitrarily complex appliance and efficiently find its activations while disaggregating. Unlike their approach, which treats each home independently, our approach leverages the ``higher order" relationship that exists between homes to disaggregate a home's total energy. Recently, Gao et al.~\cite{plaid} released a crowd-sourced data set for creating a library of appliance signatures. Like Kelly et al.~\cite{kelly2015neural}, their work only deals with appliance signatures, without considering relationship between households.

\subsection{Inter-relationship between household energy and static characteristics}
Previous work by Beckel et al.~\cite{beckel2013automatic} has shown that given aggregate energy data for a large number of homes, one can predict various household characteristics such as the family income, number of occupants, area of home. Our work is inspired from their work and shows that various household characteristics can be used to find `similarity' between homes. Additionally, many of the household characteristics that would be used by Neighborhood NILM are publicly available through city records or other sources. These values are already being used today by companies like OPOWER to match households based on similarity, for the purposes of establishing peers and norms of energy usage.

%\begin{figure}[!htb]
%	\centering
%	\includegraphics[scale=1]{figures/hvac_boxplot} 
	
%	\caption{HVAC energy consumption shows a seasonal trend. Prediction based on national average fails as it does not take into consideration such a trend and instead gives a single estimate for the entire year. Here, the green bar shows the HVAC energy proportion for Texas.}
%	\label{fig:occupancy} 
%\end{figure}

\begin{table}
	%\tabcolsep{0.3 pt}
	\begin{center}
		\begin{tabular}{p{3cm}|p{4cm}} \hline 
			\textbf{Feature category}&\textbf{Features}\\
			
			\hline
			Raw monthly energy & 12 month household energy aggregate\\
			Derived monthly energy & Variance over 12 month aggregate energy, Min. energy/Max. energy across 12 months, Max. energy-Min. energy across 12 months, (Max. energy - Min. energy)/Max. energy\\
			Static household characteristics& Area, \# occupants, \# rooms\\
			\hline 
			
		\end{tabular}
	\end{center}
	\caption{Features used in our approach for finding similarity between homes}
	\label{tab:features}
\end{table}

\section{Approach}

The goal of Neighbourhood NILM algorithm is to predict the energy consumption of
household appliances given only a single aggregate energy reading per month.
The key intuition behind Neighbourhood NILM is that `similar' homes have
`similar' energy consumption on a per-appliance basis. The basic approach is illustrated in Figure
\ref{fig:illustration}. First, we assume access to some number of homes with
direct submetering infrastructure, i.e. power meters on individual circuits or
loads. Many companies are already collected submetering information from
thousands of homes around the world. For each of these homes, we generate a set
of features, including the monthly energy bill, the size of the home, the number
of occupants, the household income, or any other features that are
available. Then, in any given test home, we generate the same feature set. We
use a K-nearest neighbor (KNN) algorithm to find $K$ neighbors for the test
home. Finally, we define the disaggregated energy usage in the test home to be
the average disaggregated energy usage in its $K$ neighbors.

This approach must address two basic questions: how to find `similar' homes, and
how to combine information from these `similar' homes to predict appliance
energy consumption for a test home. In our approach, we define ``similarity''
differently for each disaggregated load. For example, neighbors may be similar
for heating and cooling if they are in the same geographic region, and they may
be neighbors for refrigerator loads if they are roughly comparable in size. By
defining similarity differently for each load, we leverage the fact that no two
homes are exactly identical in all respects, but that every home is likely to
have a set of similar homes along any given dimension. In order to decide which
features are most indicative of similarly with respect to any given load, we use
the set of submetered homes. Since the true similarity of these homes can be
calculated based on the submetered data values, we use cross-validation to first
find the subset of home characteristics that is most correlated to similar
energy levels for a given appliance. Then, we use that set of features to find
neighbors in that dimension. The three steps to apply Neighborhood NILM are described in more detail below.

\begin{figure*}[t]
	\centering
	\includegraphics[scale=1]{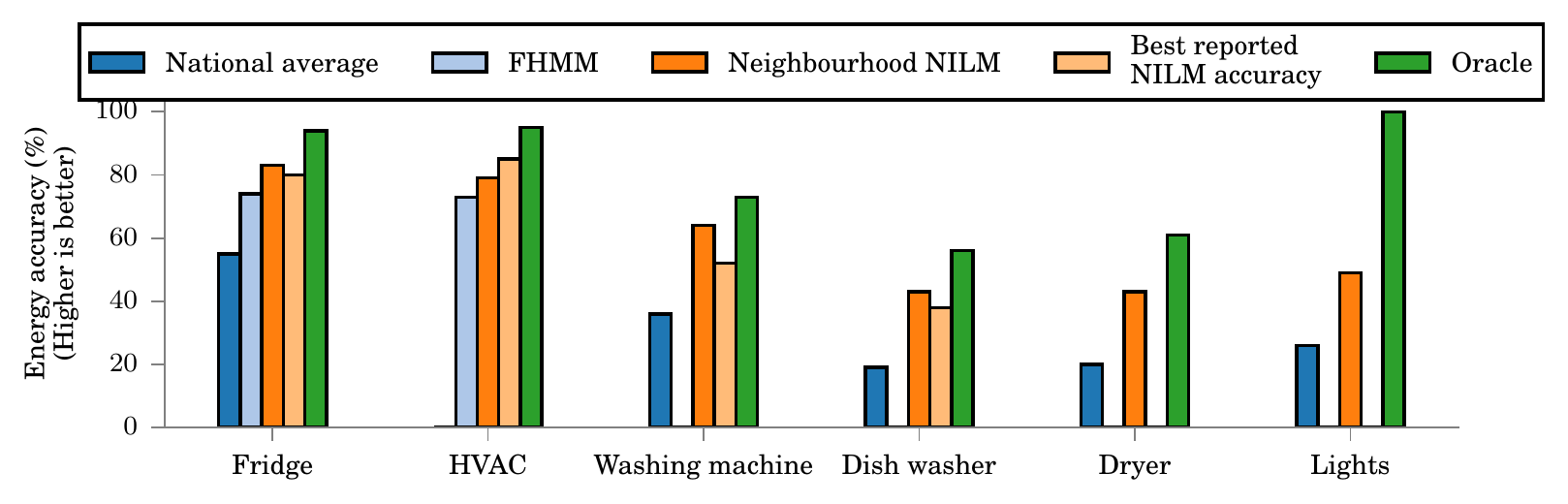} 
	
	\caption{Across all appliances, Neighbourhood NILM gives comparable or better performance than best reported NILM accuracy and better performance than all other baseline approaches. Neighbourhood NILM achieve 83\% and 79\% accuracy disaggregating fridge and heating/cooling loads, compared to 74\% and 73\% for FHMM. It also achieves upto 64\% accuracy on washing machine, dish washer, dryer and lights, all of which show poor accuracy using traditional NILM techniques.}
	\label{fig:main-result} 
\end{figure*}

\noindent \textbf{I: Feature extraction and normalisation:} In this step, we try to find candidate features for defining similarity between homes.  Historical monthly energy data can suggest seasonal trends. Homes having similar historical aggregate energy consumption have similar response to seasons and thus can be considered similar. We also derive various features from historical monthly energy data such as its variance, range (difference between maximum and minimum), ratio of minimum to maximum monthly energy consumption, to learn specific time series characteristics. Static household characteristics such as area, number of occupants and number of rooms can also be used to define similarity between homes. Static household characteristics are often publicly available and are already being used by companies to match homes based on similarity. Table \ref{tab:features} summarises the features used in our approach. Next, we normalise these features in the range 0 to 1, since these features are on different scales. We compute these normalised features across both the submetered and the test home.

\noindent \textbf{II: Finding neighbourhood:} Given a set of features, `similarity' between homes can be found using distance functions. Standard distance functions such as Euclidean, Manhattan, etc. can be used. For each test home, we find its `neighbourhood' by finding its $K$ nearest neighbours from the set of submetered homes, akin the well-known $K$ nearest neighbours algorithm.

\noindent \textbf{III: Predicting appliance energy consumption:} Having found $K$ nearest neighbours for the test home, we predict the energy consumption of an appliance in this test home as the average of energy consumption of that appliance across the $K$ submetered homes. While we use the regular average, we can also use weighted average, which assigns higher weights to closer neighbours while making the prediction. We leave using weighted average for future work.

\begin{table}
	%\tabcolsep{0.3 pt}
	\begin{center}
		\begin{tabular}{p{3cm}|p{3cm}} \hline 
			\textbf{Appliance}&\textbf{Percentage contribution to aggregate}\\ \hline
			HVAC& 13\\
			Lighting&11\\
			Refrigeration&7\\
			Dryer&4\\
			Dish washer&2\\
			Washing machine&1\\ 
			\hline 
			
		\end{tabular}
	\end{center}
	\caption{Contribution of different appliances to household aggregate across the US~\cite{eia}}
	\label{tab:national-average}
\end{table}

\section{Evaluation}

\subsection{Data set}

We use the publicly available Dataport data set~\cite{parson2015dataport} for evaluating our approach. Dataport data set contains aggregate and appliance level power information for more than 700 homes in Texas, USA for up to 3 years. Submetered and household aggregate power data was collected at 1-minute resolution. 70 homes contain data for 1 year (2013) across 6 appliances of interest (fridge, HVAC, lights, dryer, washing machine and dish washer) and house metadata such as the number of occupants, area of the home and number of rooms. While the data set contains submetered data from a large number of appliances, only the chosen 6 appliances had data across significant number of homes\footnote{\url{http://blog.oliverparson.co.uk/2014/06/wikienergy-data-set-statistics.html}}. We further filtered homes which had data collection issues and were left with 25 homes. 

\subsection{Baseline approaches and upper bound}

We compare the accuracy of our approach against the following:

\subsubsection{National average}
The US Energy Information Administration (EIA) provides a breakdown of residential electricity consumption in the US by end use~\cite{eia} as shown in Table \ref{tab:national-average}. This provides an estimate of average contribution of different appliances to the aggregate household consumption. The ``national average" energy consumption for an appliance is computed  to be the fraction of aggregate household energy, as given by national estimates in Table \ref{tab:national-average}. It must be noted that US national average is going to under estimate the HVAC energy consumption for the Texas region, especially in the summer months. Texas shows higher temperature than most US states and thus has a higher contribution of 18\% for HVAC loads, in comparison to the 13\% in the national average~\cite{eia-texas}. For HVAC, we use the Texas average, and for all other appliances (for which we don't have Texas data), we use the national average while computing energy consumed by this baseline approach.

\subsubsection{Factorial Hidden Markov Model (FHMM)}
FHMM is a well known NILM technique~\cite{kolter_2012, parson_2012}. In this approach, each appliance is modelled as a hidden Markov model (HMM) containing `n' states, where a state indicates mode of operation of the appliance. Many appliances can be modelled as 2 state appliances (ON or OFF). The HMM for each appliance is described by 3 parameters: 1) Initial probability ($\pi$)- containing the probability of each state at the initial time; 2) Transition matrix ($A$)- containing the probability of transition between states (e.g. ON to OFF ); and 3) Emission matrix ($B$)- containing the power draw distribution of different states of the appliance.

\subsubsection{Best reported accuracy}
We look at prior literature~\cite{shao, batra2015if, parson_2012, redd, nilmtk, kolter_2010, kolter_2012, parson_2014, kelly2015neural} in top venues pertaining to energy disaggregation research and present the best reported accuracy for each appliance. This represents the best reported accuracy thus far, with the caveat that these numbers may be obtained under varied experimental conditions (assumptions, data set, etc.). It must be pointed that the state-of-the-art shows poor accuracy across low power loads such as lights, and complex loads such as washing machine and dish washer.

\begin{figure*}[t]
	\centering
	\includegraphics[scale=1]{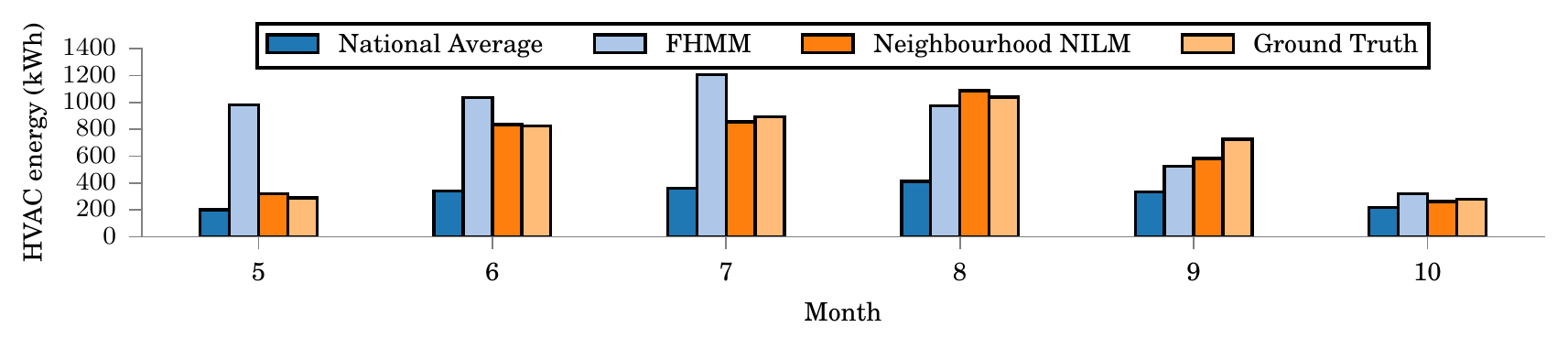} 
	
	\caption{Neighbourhood NILM captures the trend in HVAC energy consumption across the months, which none of the other approaches does.}
	\label{fig:occupancy} 
\end{figure*}

\subsubsection{Oracle}
For any test home, the oracle finds the `optimal' subset of `neighbourhood' homes from all possible sub sets of homes for each home for each load. The appliance energy data from these `optimal' homes is then averaged to predict the appliance energy consumption for the test home. Here, `optimality' is defined as giving the best energy disaggregation accuracy.  The oracle represents the upper bound on the accuracy obtainable by our approach, under the setting when there exists an optimal distance function, feature set and number of neighbours which can find the optimal `neighbourhood' of homes for each test home. 

\subsection{Evaluation metric}
Based on the metrics used in the prior literature~\cite{nilmtk, kolter_2012}, we define Energy accuracy (\%) as:
\vspace{2mm}
\\ 100 - $\frac{|\text{Predicted appliance energy - Actual appliance energy}|\times 100\%}{\text{Actual appliance energy}}$

Given that the ratio $\frac{|\text{Predicted appliance energy - Actual appliance energy}|}{\text{Actual appliance energy}}$ can be more than 1, when the prediction is way off from the actual usage, accuracy in such cases is considered to be 0. Higher `Energy accuracy (\%)' indicates better disaggregation performance.

\subsection{Experimental setup}
Our experimental setup tries to replicate the following real world scenario- we have a small subset of submetered homes and a large number of homes without smart meters. However, our data set has the limitation that only a small number of homes (25) containing aggregate and appliance energy for long duration (1 year) and household characteristics is available. Thus, we use the \textit{Leave-one-out} cross validation technique to evaluate our approach, where we assume the set of submetered homes to be all the homes except the test home. 

We use data from 2013 for our evaluation. For all homes, we had aggregate and appliance monthly energy consumption for the 12 months. Our task was to predict the monthly energy consumption for each appliance across these 25 homes. For the HVAC, the evaluation was done only on the months (May-October) in which it is typically used in Texas. For our approach, we do a sweep on the number of neighbours ($K$) and all subsets of features, for all appliances, and report the best accuracy. This sweep is done on a per-appliance basis, denoting that each appliance may have a different feature for `similatity' between homes.

To compare our performance with FHMM, we train an FHMM on the same home as the one on which we test using the NILMTK implementation~\cite{nilmtk, kelly2014nilmtk}. Since, FHMM is exponential in the number of appliances modelled, we choose top-5 appliance across each home, as is the practice in the NILM community~\cite{batra2015if}. Further, we modelled each appliance to consist of two states (ON or OFF). The FHMM training was done on data from August when all appliances were actively used. It must be noted that the result of FHMM disaggregation is a 1-minute power signal for each appliance. We sum up this signal for all months to get the FHMM monthly energy prediction.

\subsection{Results}
Our main result in Figure \ref{fig:main-result} shows that Neighbourhood NILM gives comparable or better disaggregation performance across all loads in comparison to FHMM, estimates from national average and best reported NILM accuracy. The optimal combination of $K$ neighbours and feature subsets is shown in Table \ref{tab:top-features}. The fact that the `Oracle' achieves sufficiently higher accuracy than our approach indicates that there is a scope to further improve our results using better features. Further, the `Oracle' establishes our approach to be promising given that it outperforms best reported accuracy in NILM literature. The disaggregation performance of FHMM on washing machine, dish washer, dryer and lights is poor and their accuracy is 0\%. This is in consonance with previous literature showing that these loads are particularly hard to disaggregate~\cite{barker_2013}. In contrast, our approach gives accuracy upto 64\% for these appliances.

\begin{table}
	%\tabcolsep{0.3 pt}
	\begin{center}
		\begin{tabular}{p{1cm}|c|p{4cm}} \hline 
			\textbf{Appliance}&\textbf{Optimal \# neighbours}&\textbf{Optimal features}\\
			
			\hline
			Fridge& 3& Raw monthly energy, Derived monthly energy, \#rooms \\
			HVAC & 5 & Raw monthly energy\\
			Washing machine & 5 & Raw monthly energy, Derived monthly energy, \#occupants \\
			Dish washer & 7 & \#occupants, area\\
			Dryer & 1 & Raw monthly energy, Derived monthly energy, \#rooms, \#occupants \\
			Lights & 2 & Raw monthly energy, area \\

			\hline 
			
		\end{tabular}
	\end{center}
	\caption{Optimal features and number of neighbours for different household appliances}
	\label{tab:top-features}
\end{table}

Neighbourhood NILM accurately predicts the HVAC consumption and captures the seasonal trend (Figure \ref{fig:occupancy}). The prediction based on national average fares poorly. This can be explained by the fact the national average numbers are computed on the entire year and not on a per-month basis and thus show poor accuracy for appliances such as HVAC which have heavy seasonal variations. In general, FHMM gives comparable accuracy. However, its accuracy in the May (5th month) is poor, when it over predicts HVAC energy consumption. This is likely due to the presence of a load having similar power characteristics such as the HVAC. Our approach is agnostic of such load profiles.

\begin{figure*}[t]
	\centering
	\includegraphics[scale=1]{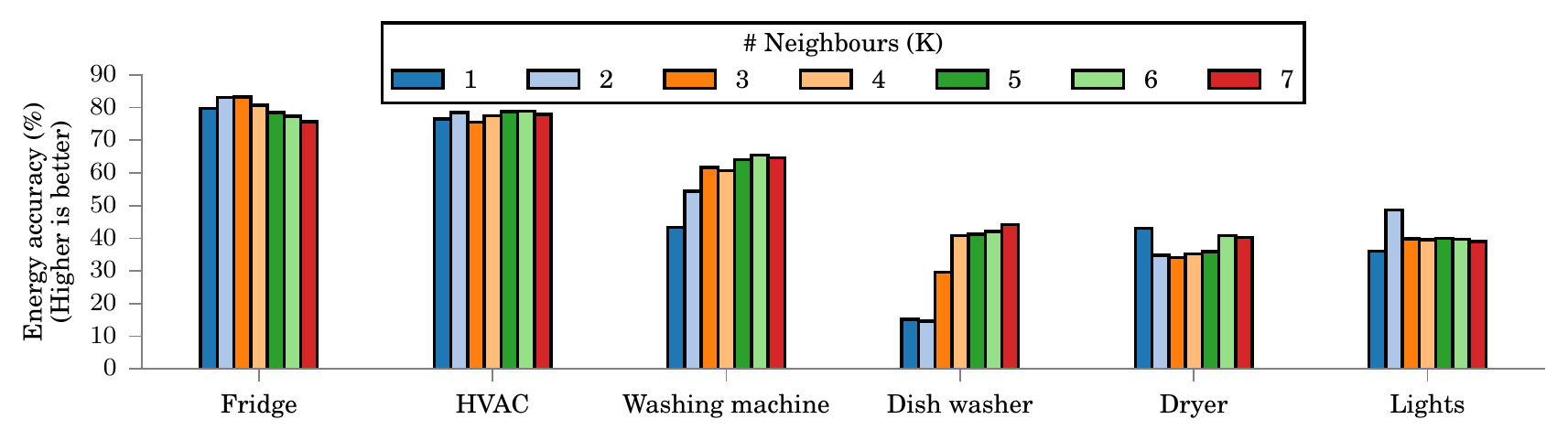} 
	
	\caption{With the exception of dryer and lights, our approach achieves the best accuracy for $K$=3 to 6 neighbours. Heavily used appliances such as fridge and HVAC show little variation in accuracy in comparison to other appliances that possibly have huge variance across homes and thus their accuracy varies significantly with $K$.}
	\label{fig:sensitivity-k} 
\end{figure*}

\subsection{Sensitivity analysis}
Having described the `highest' accuracy achievable by our approach for all appliances, we now do a sensitivity analysis across the number of neighbours ($K$) and subsets of feature vectors. While computing the sensitivity of our approach to $K$, we choose the optimal feature (from Table \ref{tab:top-features}) for each appliance and vary $K$. Figure \ref{fig:sensitivity-k} shows the variation in accuracy across different $K$. Heavily used appliances such as fridge and HVAC show little variation in accuracy with varying $K$. Other appliances have a huge variance across homes and thus their accuracy varies significantly with $K$. The dish washer in particular has low accuracy for small number of neighbours. Since the prediction based on a larger number of neighbours would be a weighted factor of prediction based on smaller number of neighbours, dish washer shows low accuracy. This indicates that the feature set for dish washer needs further work.

\begin{figure*}[t]
	\centering
	\includegraphics[scale=1]{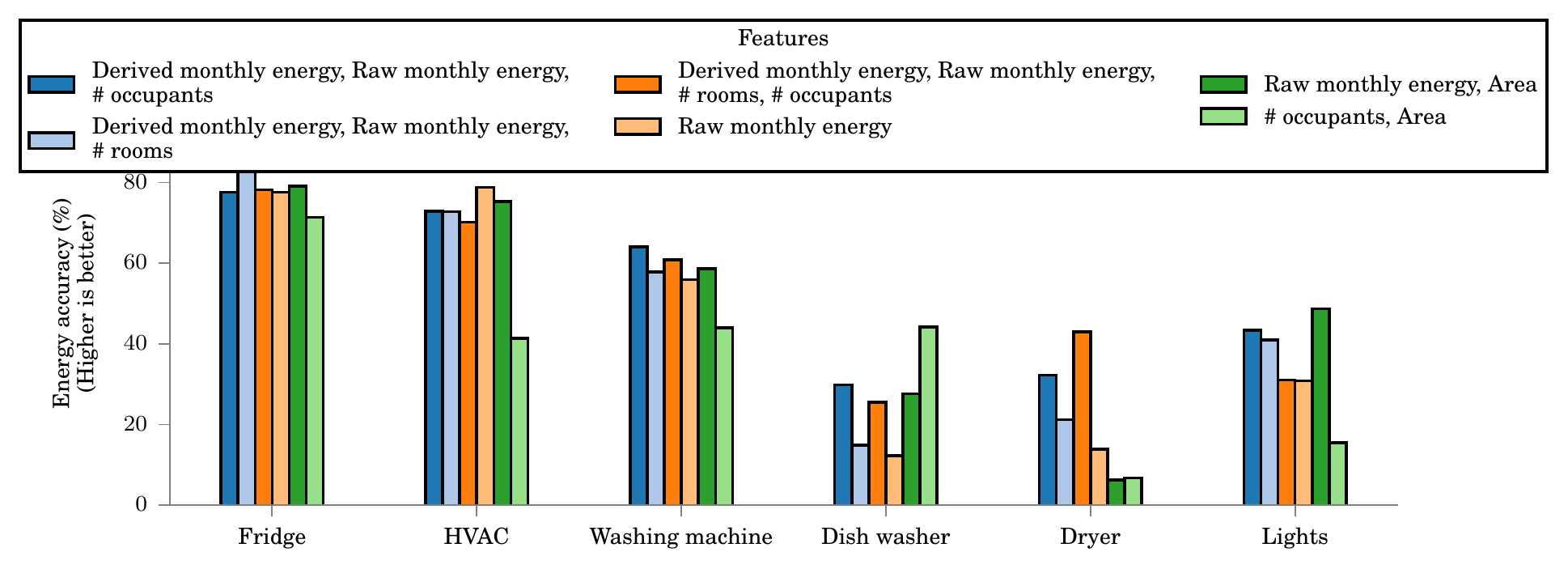} 
	
	\caption{Washing machine, dish washer and dryer achieve best accuracy when \# occupants is one of the features possibly explained by the relationship between \# occupants and \# dishes and \# clothes; for HVAC, the raw monthly energy feature gives best accuracy as it accurately captures the seasonal trends; for lights, area is an important feature possibly explained the fact that larger homes need more lighting fixtures. }
	\label{fig:sensitivity-features} 
\end{figure*}

While computing the sensitivity of our approach to features, we choose the optimal $K$ (from Table \ref{tab:top-features}) and test our approach on different features. There is no single feature which best describes the `similarity' across all appliances as shown in Figure\ref{fig:sensitivity-features}. For washing machine, dish washer and dryer; number of occupants is one of the features in the optimal feature set. This dependence can be explained by the fact that there is usually a correlation between the number of people and number of clothes and dishes. For the HVAC, historical monthly energy trend is the best feature, which can be explained by the fact that the energy trend best captures the seasonal variations, which play a major role in HVAC energy consumption. Household area features helps improve the disaggregation accuracy of lights. This can be explained by the fact that larger homes in general have more lighting fixtures.

\section{Limitations and Future work}
Our work has five major limitations, described below:

\noindent\textbf{1:}Our work has only been tested on a single data set which had homes from a single geographical location (Texas), which are related being from the dataset. While many NILM data sets have been released in the past, none of the other data set satisfied the requirements of containing both aggregate and submetered power data for a large number of homes for at least one year. While the HES data set~\cite{hes} contains data from a large number of homes for a large time, it does not contain aggregate power data. Given the high variability present in submetering across homes, we can't assume the aggregate to be the sum of submetered loads. The ECO data set~\cite{eco} contains data from 6 homes for 8 months. But, there is no single load which is submetered across all these 6 homes. AMPds~\cite{ampds}, iAWE~\cite{iawe}, Blued~\cite{blued} contain data only from a single home. REDD~\cite{redd} contains data for a very short duration from 6 homes. While UK-Dale~\cite{UK-DALE} contains data from 6 homes, only one of them has year long data. We believe that this concern can be addressed by considering geographical co-ordinates as a feature while computing similarity between homes.

\noindent\textbf{2:}Furthermore, if people respond to energy breakdown and change their energy consumption, our feature vector consisting of historical energy usage is likely to be less effective. We believe that such changes can be captured by running changepoint detection algorithms over longer duration historical energy data. 

\noindent\textbf{3:}Since our approach only uses single reading per month, it can only estimate the monthly energy consumed by individual appliances, and cannot estimate the finer grained power signal. Estimating the power signal has been shown to enable various applications~\cite{batra2015if, alcala2015detecting}. However, in homes that do have AMI metering infrastructure, the results of Neighborhood NILM could potentially improve. This is a direction of future work.

\noindent\textbf{4:}`Outlier' homes (such as homes consuming very high energy), which are likely candidates for energy feedback may have wrong predictions owing to the fact that `neighbouring' homes are not representative of their energy consumption. We plan to address this concern in future work by first segregating these `outlier' homes based on a model of home's energy consumption as a factor of various static and dynamic characteristics such as: area, number of occupants, etc.

\noindent\textbf{5:}This approach is less likely to work for very uncommon appliances. It is likely limited to appliances which are commonly present across a large number of homes. This concern is common to current NILM algorithms as well, which are typically suited for more common appliances. The generality of this approach will grow linearly with the number of homes on which submetering infrastructure is deployed.

%	\begin{figure}[!htb]
%		\centering
%		\includegraphics[scale=1]{figures/eco_vs_wiki} 
%		
%		\caption{Energy consumption patterns can vary widely across geographies (red lines represent data from ECO dataset collected in Switzerland (~\cite{eco}), black lines represent data from Dataport dataset ). Thus, our approach is likely applicable only within a geographical region. }
%		\label{fig:eco-vs-wiki	} 
%	\end{figure}
	
\section{Conclusions}
For disaggregation purposes, traditional NILM techniques only use data from the home where they are disaggregating. Further, traditional NILM techniques require a smart meter per home. In this paper, we presented a novel technique which finds homes `similar' to the test home and combine information from them to predict appliance energy consumption. Our approach works when a single reading is available per month (as is the case with traditional electricity meters). Results show that our approach outperforms state-of-the-art NILM technique which rely on smart meters. Neighbourhood NILM can achieve 83\% and 79\% accuracy disaggregating fridge and heating/cooling loads, compared to 74\% and 73\% for a technique called FHMM. Furthermore, it achieves up to 64\% accuracy on washing machine, dryer, dishwasher, and lighting loads, which is higher than previously reported results. Many existing techniques are not able to disaggregate these loads at all. These results indicate a potentially substantial advantage to installing submetering infrastructure in a select few homes rather than installing new high-frequency smart metering infrastructure in all homes.

%\scriptsize
\bibliographystyle{abbrv}
\bibliography{reference}  % sigproc.bib is the name of the Bibliography in this case

\ifCLASSOPTIONcaptionsoff
  \newpage
\fi

% trigger a \newpage just before the given reference
% number - used to balance the columns on the last page
% adjust value as needed - may need to be readjusted if
% the document is modified later
%\IEEEtriggeratref{8}
% The "triggered" command can be changed if desired:
%\IEEEtriggercmd{\enlargethispage{-5in}}

% references section

% can use a bibliography generated by BibTeX as a .bbl file
% BibTeX documentation can be easily obtained at:
% http://www.ctan.org/tex-archive/biblio/bibtex/contrib/doc/
% The IEEEtran BibTeX style support page is at:
% http://www.michaelshell.org/tex/ieeetran/bibtex/
%\bibliographystyle{IEEEtran}
% argument is your BibTeX string definitions and bibliography database(s)
%\bibliography{IEEEabrv,../bib/paper}
%
% <OR> manually copy in the resultant .bbl file
% set second argument of \begin to the number of references
% (used to reserve space for the reference number labels box)
%for a photo:

\end{document}